\documentclass{bmvc2k}

\usepackage{graphicx}
\usepackage{amsmath}
\usepackage{amssymb}
\usepackage{bm}
\usepackage{textcomp} 
\usepackage[linesnumbered,ruled,vlined]{algorithm2e}
\usepackage{algpseudocode}
\usepackage{hyperref}

\title{Semi-supervised Active Learning for Instance Segmentation via Scoring Predictions}

\addauthor{Jun Wang\textsuperscript{$\dagger$}}{wangjun916@pingan.com.cn}{1}
\addauthor{Shaoguo Wen\textsuperscript{$\dagger$}}{wenshaoguo0611@gmail.com}{12}
\addauthor{Kaixing Chen}{chenkaixing630@pingan.com.cn}{1}
\addauthor{Jianghua Yu}{yujianghua603@pingan.com.cn}{1}
\addauthor{Xin Zhou}{zhouxin879@pingan.com.cn}{1}
\addauthor{Peng Gao}{gaopeng712@pingan.com.cn}{1}
\addauthor{Changsheng Li (Corresponding Author)}{lichangsheng507@gmail.com}{3}
\addauthor{Guotong Xie (Corresponding Author)}{xieguotong@pingan.com.cn}{145}

\addinstitution{
 Ping An Healthcare Technology,\\
 Beijing, China
}
\addinstitution{
 School of Information and Communication Engineering,\\
 Beijing University of Posts and Telecommunications, \\
 Beijing, China
}
\addinstitution{
 School of Computer Science and Technology,\\
 Beijing Institute of Technology, \\
 Beijing, China
}

\addinstitution{
Ping An Health Cloud Company Limited., \\
Shenzhen, China
}

\addinstitution{
Ping An International Smart City Technology Co., Ltd.,\\
Shenzhen, China
}

\runninghead{Wang et al.}{Active Learning for Instance Segmentation}

\begin{document}

 
\maketitle

\begin{abstract}
Active learning generally involves querying the most representative samples for human labeling, which has been widely studied in many fields such as image classification and object detection. However, its potential has not been explored in the more complex instance segmentation task that usually has relatively higher annotation cost. In this paper, we propose a novel and principled semi-supervised active learning framework for instance segmentation. Specifically, we present an uncertainty sampling strategy named Triplet Scoring Predictions (TSP) to explicitly incorporate samples ranking clues from classes, bounding boxes and masks. 
Moreover, we devise a progressive pseudo labeling regime using the above TSP in semi-supervised manner, it can leverage
both the labeled and unlabeled data to minimize labeling effort while maximize performance of instance segmentation. Results on medical images datasets demonstrate that the proposed method results in the embodiment of knowledge from available data in a meaningful way. The extensive quantitatively and qualitatively experiments show that, our method can yield the best-performing model with notable less annotation costs, compared with state-of-the-arts.

\end{abstract}

\section{Introduction}
\label{sec:intro}
State of the art deep neural networks have shown favorable performance in instance segmentation tasks~\cite{he2017mask,munjal2020towards}, but training these supervised deep instance segmentation models involves labeling a large scale fine-grained dataset~\cite{majumder2019content}, which is much more time-consuming and costly to obtain than classification~\cite{wang2016cost, sensoy2018evidential} and object detection~\cite{li2013adaptive} tasks. An ideal architecture would integrate data labeling and model training in a meaningful way, so as to maximize model performance with minimal amount of labeled data. Fortunately, active learning (AL) provides a machine learning paradigm to mitigate this burden. It promises to help reduce the efforts of data annotation~\cite{settles2009active,settles2011theories,wang2016cost}, through intelligently selects a subset of informative samples from a large unlabeled data pool. 
In the past decades, lots of active learning approaches have been proposed~\cite{settles2009active,settles2011theories,wang2016cost}, and many have been successfully applied to image classification~\cite{joshi2012scalable, joshi2009multi, li2013adaptive, wang2016cost,smailagic2018medal}, object detection~\cite{li2013adaptive, desai2019adaptive,kao2018localization,roy2018deep}, semantic segmentation~\cite{Blanch2017ActiveDL, Yang2017SuggestiveAA, Scandalea2019DeepAL}, etc. 
Unfortunately, researchers pay little attention to AL for instance segmentation task until now. Meanwhile, most works on active learning, especially deep learning based methods, usually ignore the progress of the closely related field like semi-supervised learning (SSL)~\cite{arazo2019pseudo,tarvainen2017mean,bortsova2019semi,albalate2013semi}, that can effectively take advantage of the left unlabeled samples. In particular, we can regard those samples with high
prediction confidence as pseudo labeled data, and add them into the training set for model training without extra
human labor. More importantly, in order to lower the noisy effect of pseudo label, we iteratively update the pseudo label set based on current updated model. Besides, generally object segmentation in medical images is much more difficult than that in natural images~\cite{hao2012combining}, due to the poor image quality with low contrast, heavy speckle noise, large variation of lesion in poorly defined shape and appearance, especially between the benign and the malignant, thus, medical image datasets are used to verify the effectiveness of proposed method. 

{\bf Contributions:} Our method significantly differs from the existing works in the following aspects: (1) To the best of our knowledge, our approach constitutes the first attempt to actively query informative samples for instance segmentation in a semi-supervised setting. (2) Considering that an AL framework for instance segmentation should
 take class, bounding box and mask into consideration simultaneously, we introduce a Triplet Scoring Predictions (TSP) of both classification, masks and bounding boxes branches, which can offer a more reasonable measure of sample's uncertainty for boosting model performance. (3) We adopt a novel label-efficient pseudo labeling strategy based on the above triplet scoring prediction, in a semi-supervised manner. A novel loss is designed to lower the noisy effect of pseudo label. Consequently, our architecture can allow exploiting more reasonable sampling criteria and adaptive pseudo-labeling loss to obtain significant savings in annotation cost, compared with state-of-the-arts.

\vspace{-15pt}

\section{Related Work}

As there are few published papers on AL for instance segmentation, thus, our method is most related to AL for semantic segmentation task. Besides, both AL and SSL aim to improve learning with limited labeled data, thus they are naturally related. Then, we also briefly review the related semi-supervised pseudo labeling work for AL, which is proved as the most effective and concise treatment.

{\bf AL for semantic segmentation:} Regarding to AL for semantic segmentation, samples are usually selected based on the inconsistency of model output predictions at different MC Dropout conditions~\cite{gorriz2017cost, ozdemir2019active}, or using Core set method~\cite{sener2017active} to choose a subset, such that the largest distance between chosen point and unlabeled points is minimized in the feature space. Noteworthily, most of the basic AL methods usually fail to use information from inner layers of DNN.
On the other hand, The network-agnostic learning-based approaches like Learning Loss~\cite{yoo2019learning} uses an auxiliary network module and loss function to learn a measure of information gain from new samples. However, it is also restrictedly validated on classification, human pose estimation and object detection task with simple SSD~\cite{liu2016ssd}. Thus, to the best of our knowledge, the most relevant work to ours comes from Mask Scoring RCNN~\cite{huang2019mask}, which also proposed a network block to learn the quality of the predicted instance masks. The difference between our work and Mask Scoring RCNN are two-folds. First, the motivation is different, Mask Scoring RCNN improves instance segmentation performance by prioritizing more accurate mask predictions, our work scores the predictions for defining a more reasonable uncertainty strategy in active learning for instance segmentation. Second, the detailed architecture of the new branch is different, in addition, we synthetically predict the bounding box score except for predicting the mask score.

\noindent{\bf Semi-supervised pseudo labeling:} The pseudo labeling is proved as a simple and effective method in semi-supervised learning~\cite{lee2013pseudo, zhu2016multi, bortsova2019semi, albalate2013semi}, which generates pseudo label for unlabeled data using model's certain predictions. However, given that the naive pseudo-labeling overfits to incorrect pseudo-labels due to the so-called confirmation bias~\cite{arazo2019pseudo, tarvainen2017mean}, which would damage the performance of model. Unfortunately, this situation could get worse when applying pseudo labeling to the instance segmentation. Specifically, If we only take classification score to construct thresholds for pseudo labeling, there will be many false-positive instances on the pseudo-labeling data set. Considering that most works in pseudo labeling usually focused on the image classification~\cite{wang2016cost}, therefore, this paper constitutes attempt to pseudo-labeling for instance segmentation from the network predictions. 

\begin{figure}

\centering
\includegraphics[width=5in]{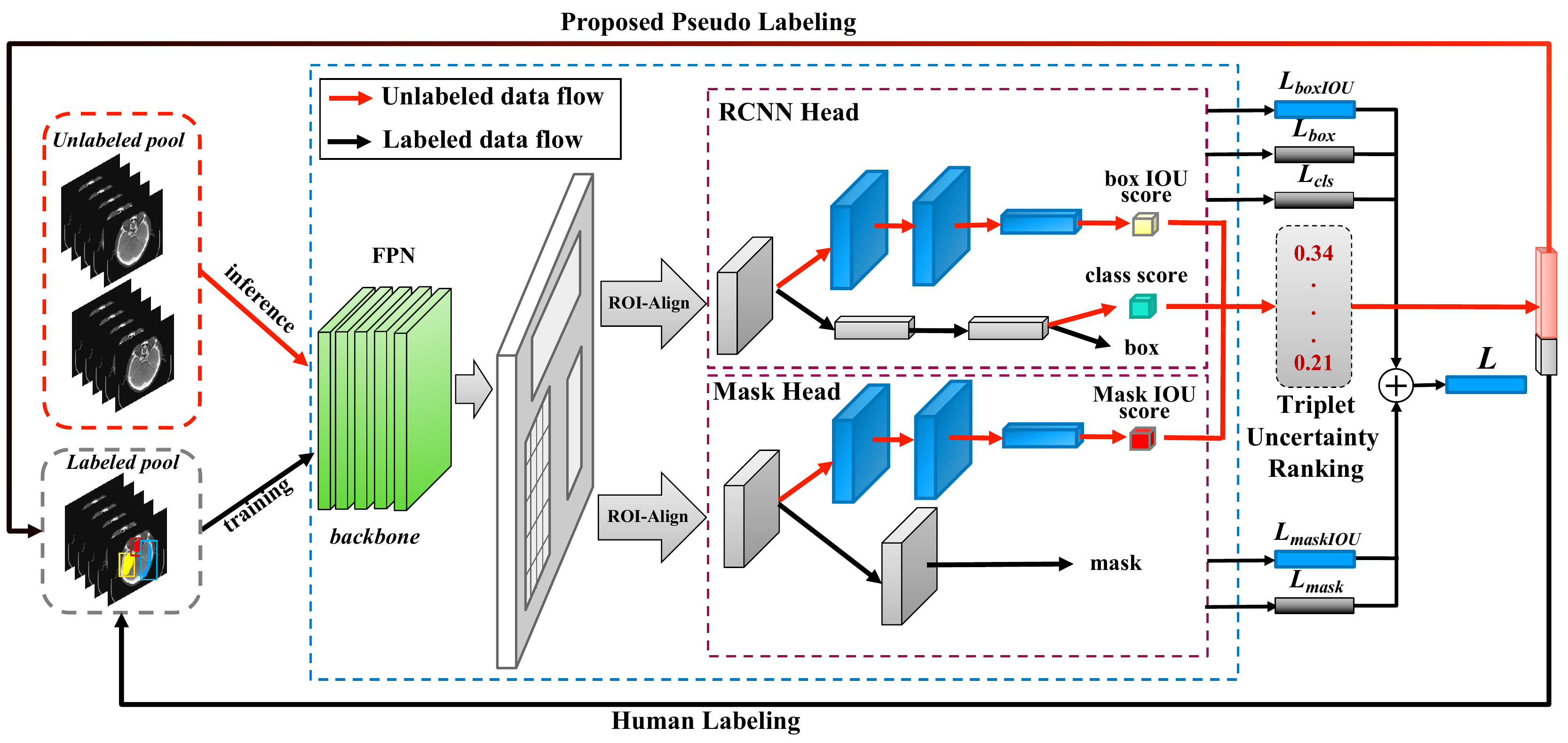}
\caption{Illustration of the overall architecture. Two new parallel branches  (in blue color) are attached to the standard RCNN head and Mask head. The output of each branch is a scalar value which predicts the quality of predicted bounding box and mask of each instance respectively. The triplet scoring predictions (TSP) of each instance are obtained by combining these two scores and the classification score, which is used for the proposed pseudo labeling(upper arrow) and active learning(bottom arrow). The model is trained on the labeled pool which contains labeled data and pseudo-labeled data. Noting that, the black lines represent the training flow, red lines represent the inferencing flow.} 

\label{fig1}
\end{figure}


\section{Methodology}

An overview of the proposed architecture is illustrated in Figure 1. Our semi-supervised active learning framework mainly consists of the following two components: (1) A triplet scoring prediction (TSP) explicitly gathering the ranking clues from the classification, the bounding box and the mask branches. (2) A progressive pseudo labeling regime using the above triplet scores to leverage both the labeled and unlabeled data. Details can be found in following sections.

Firstly, triplet scoring prediction (TSP) is proposed to explicitly measure the integrated uncertainty from the classification, the bounding box and the mask. To this end, two novel parallel branches  (in blue color) are attached to the standard RCNN head and Mask head separately. Each branch has two convolution layers(all have kernel=3, stride=2 and padding=1) and  a global average pooling layer followed by one fully connected layer. More importantly, the output of each branch is a scalar value after being scaled by a sigmoid activation function, which predicts the IoU of predicted bounding box and target bounding box in RCNN head and the IoU of predicted mask and target mask in Mask head respectively. 

Secondly, the above triplet scores are used for {\bf active learning(bottom arrow)} and the {\bf pseudo labeling(upper arrow)}. In particular, informative samples are selected for manual annotation by ranking of the triplet scoring. On the other hand, according to the proposed pseudo labeling strategy, the rest of the unlabeled samples are pseudo labeled according to the predictions of the trained model in previous cycle, and added into the train set to train the model in current cycle. Note that each instance on the pseudo-labeling samples is also selected by setting proper thresholds via the aforementioned triplet scoring. Moreover, a novel loss strategy is designed for handling the false negative on the pseudo-labeling phase. Consequently, the model is trained on the labeled pool which contains labeled data and pseudo-labeled data, with loss consisted of 5 branch losses from supervised aspect and an implicit pseudo labeling loss from semi-supervised aspect.

\subsection{Novel Uncertainty Strategy via Triplet Scoring Predictions}

The uncertainty score plays an important role in active learning~\cite{settles2009active,settles2011theories,zhou2017fine}, which is utilized as ranking clues for selecting the informative samples. The entropy of classification prediction is usually used as
the uncertainty score in the traditional active learning schemes~\cite{wang2016cost}. Unfortunately, it is mainly designed for the classification task, other than the best choice for the instance segmentation task, since it cannot fully measure the quality of instance localization and instance masking. To this end, we constitute the first attempt to explicitly formulate more reasonable uncertainty definition to apply active learning for instance segmentation. 

For the instance segmentation task, which solves both object detection and semantic segmentation, locating object instances with pixel-level accuracy, as a result it requires much higher labeling cost~\cite{he2017mask}. As is mentioned above, since instance segmentation can predict the classification score, bounding box and mask for each instance, but the commonly used classification score~\cite{wang2016cost} cannot comprehensively reflect the quality of the predicted bounding box and mask, so it is insufficient to adopt only classification score to measure uncertainty for instance segmentation task in active learning. Furthermore, even if the bounding box and mask can be predicted by the model, there are no published works simultaneously measure the quality of the predicted bounding box and mask on the unlabeled samples. Therefore, we propose a novel architecture to obtain the triplet scores of the predicted class probability, bounding box and mask directly, which doesn't require any ground truth when inferencing on the unlabeled data set.

{\bf Training (Black line flow in Figure 1):} The entire model integrated with the mask IoU head and the bounding box IoU head is trained in an end-to-end fashion and doesn't require extra labels. The input of the bounding box IoU head is the features of positive ROIs extracted by RoI-Align, here positive ROIs mean that the proposal ROIs generated by RPN which have a IoU with the matched ground truth box larger than 0.5, and the IoU of predicted bounding box and matched ground truth box is used as the target of bounding box head. Similarly, The input of the mask IoU head is same as the input of mask head for predicting instance mask. In particular, we binarize the predicted mask by setting a threshold of 0.3, the IoU of binarized mask and matched ground truth mask is used as the target of mask IoU head. $\textit{l}_2$ loss is applied on both bounding box head and mask IoU head for training, then the total loss of model is define as:

\vspace{-10pt}

\begin{equation}
\begin{aligned}
L &= \frac{1}{N_{cls}}\sum_{i}L_{cls}(p_i, p_i^{*})+\frac{1}{N_{box}}\sum_{i}p_i^{*}L_{box}(t_i, t_i^{*})+\frac{1}{N_{mask}}\sum_{i}p_i^{*}L_{mask}(m_i, m_i^{*})\\
&+\lambda\frac{1}{N_{box}}\sum_{i}p_i^{*}\textit{l}_2(biou_i, biou_i^{*})+\lambda\frac{1}{N_{mask}}\sum_{i}p_i^{*}\textit{l}_2(miou_i, miou_i^{*}).
\end{aligned}
\end{equation}
Where the ground-truth label $p_i^{*}$ is 1 if the anchor is positive, and is 0 if the anchor is negative. $\lambda$ is the loss weight. More detained about loss function can be found in Faster RCNN ~\cite{ren2015faster} and Mask RCNN~\cite{he2017mask}.

{\bf Inference (Red line flow in Figure 1):} In order to reduce the cost of computation, NMS is applied to detection results to get the final predicted bounding box during inference, then the feature of the final predicted bounding box extracted by RoI-Align is feed in the bounding box head and mask head to get the bounding box score and mask score respectively, note that there is less ROIs-feature feed in the bounding box head and mask head comparing with the training period.

{\bf Novel Uncertainty Strategy:} Our proposed framework can predict the class score $c_i^j$, bounding box score $b_i^j$ and mask score $m_i^j$ for the $j$th instance on the $i$th image when inferencing on the unlabeled data set. We define the uncertainty score from two aspects: the informativeness and diversity of the triplet scores. The informativeness is measured by the mean of these devised uncertainty scores. On the other hand, the diversity is defined by the standard deviation of these three branches, more precisely, the difference between these three branches indicates that the amount of information carried on this instance, since the instance should be more informative if it cannot maintain the consistency on these triplet scores. Then the triplet uncertainty score of the $j$th instance on the $i$th image is defined as:
\begin{equation}
s_i^j = e^{-std(c_i^j,b_i^j,m_i^j)}*mean(c_i^j,b_i^j,m_i^j)
\end{equation}
Instance-based uncertainty score can be calculated by the above equation, similarly, the image-based uncertainty score of the $i$th image is defined as:
\begin{equation}
S_i = e^{-std(\textbf{s}_i)}*mean(\textbf{s}_i)
\end{equation}
where $\textbf{s}_i$ is a vector which consists of all instance-based uncertainty scores of the $i$th image. The lower the triplet scoring, the higher the uncertainty.

\subsection{Pseudo Labeling based on Triplet Scoring Predictions}

Naturally, based on aforementioned triplet scoring, one instance can be picked up for pseudo labeling when its classification score larger than $\sigma_c$ and its bounding box score larger than $\sigma_b$ and its mask score larger than $\sigma_m$. However, there is another problem subsequently: \textit{we cannot ensure that all the instances are pseudo-labeled on an image}. On one hand, we should note that instance segmentation model is not so powerful to detect all the instances for every image. On the other hand, many predicted instances with low confidence will be removed according to active selection strategy. To deal with this issue, we optimize the loss of the pseudo-labeling data, actually, the losses for bounding box regression and mask predictions don't need to be changed because they only work on the positive RoIs generated by RPN, but the classification losses in RPN head and RCNN head should change to only classify the positive RoIs, otherwise, there may be many areas with objects on a image to be regarded as background. Therefore, the loss of the pseudo-labeling sample is rewritten as:
\begin{equation}
\begin{aligned}
L_{semi} &= \frac{1}{N_{cls}}\sum_{i}p_i^{*}L_{cls}(p_i, p_i^{*})+\frac{1}{N_{box}}\sum_{i}p_i^{*}L_{box}(t_i, t_i^{*})&+\frac{1}{N_{mask}}\sum_{i}p_i^{*}L_{mask}(m_i, m_i^{*}).
\end{aligned}
\end{equation}
We multiply $p_i^{*}$ to the first term $L_{cls}(p_i, p_i^{*})$ for only classifying the positive RoIs. Note that the losses of bounding box IoU head and mask IoU head are removed for the pseudo-labeling data, because the pseudo label is not as accuracy as manual label, it will inevitably add noises to the model and have a bad influence for predicting the quality of predicted bounding boxes and masks.
Then total loss of the model trained on labeled data and pseudo-labeling data is defined as:
\begin{equation}
\begin{aligned}
L = L_{sup} + \beta*L_{semi}.
\end{aligned}
\end{equation}
Where $L_{sup}$ is the loss for labeled data which is provided at equation(2). $\beta$ is used as a coefficient for balancing losses.

\subsection{Semi-supervised Active Learning via Scoring Predictions}
In summary, our framework of semi-supervised active learning for instance segmentation combines the novel triplet scoring uncertainty formulation and the devised pseudo labeling strategy as introduced above. 
We use the dataset $\emph{D}_{train}=\emph{D}_{al}+\emph{D}_{semi}$ to train the model in this cycle, the loss function is provided at equation(5). Our overall methodology is summarized below in Algorithm 1.

\vspace{-10pt}


\section{Experiments}
\subsection{Datasets}


To verify the effectiveness of our method, we perform the experiments on two substantially large medical image datasets: (1). {\bf The head CT scans dataset for intracranial hemorrhage (ICH) segmentation}. It contains ICH and nine types, including ncx, px, zw, yx, yw, ncls, prns, gh and midline shift, each type was annotated by senior radiologists. It is split into train dataset and test dataset, the train dataset includes 10000 pictures in total, while the test dataset has 4154 pictures in total. The images are resized by $512\times512$. (2). {\bf The Retinal OCT dataset for Edema Lesions segmentation}. The dataset is sourced from public AI Challenger 2018 competition (\url{https://challenger.ai/competition/fl2018}). Pixel-level annotations are performed on the retinal edema area (REA), pigment epithelial detachment (PED) and subretinal fluid (SRF) lesion areas. It includes 12800 OCT images with a resolution of $512\times1024$, of which the training set contains 8960 images, and the validation and test sets both contain 1920 images. The random horizontal flip is used as the online data augmentation policy for both dataset. 

\begin{algorithm}
\caption{Semi-supervised Active Learning for Instance Segmentation}
 \textbf{initialization:}
 unlabeled dataset $\emph{D}_u$, labeled dataset for active learning $\emph{D}_{al}=\emptyset$, pseudo-labeled dataset $\emph{D}_{semi}=\emptyset$, number of selected images in each cycle $b$, maximum number of cycles $K$.

\For{$k=1$ to $K$}
{
    \If{$k==1$}
    {
        randomly selecting $b$ images to $\emph{D}_{al}$, $\emph{D}_u=\emph{D}_u-\emph{D}_{al}$;\\
        using $\emph{D}_{al}$ to train model and get trained model $M_k$;\\
    }
    \Else
    {
        using $M_{k-1}$ to predict every image in $\emph{D}_u$;\\
        calculating uncertainty score for every image in $\emph{D}_u$ according to equation(3) and get uncertainty score vector $\textbf{S}$;\\
        ranking $\textbf{S}$ and selecting top $b$ images added into $\emph{D}_{al}$, $\emph{D}_u=\emph{D}_u-\emph{D}_{al}$;\\
        letting $\emph{D}_{semi}=\emph{D}_u$, removing the $j$th predicted instance of the $i$th image in $\emph{D}_{semi}$ whose $c_i^j<\sigma_c$ and $b_i^j<\sigma_b$ and $m_i^j<\sigma_m$. The left instances in $\emph{D}_{semi}$ are pseudo-labeled;\\
        using $\emph{D}_{train} = \emph{D}_{semi}+\emph{D}_{al}$ to train model and get trained model $M_k$;
    }
}

\end{algorithm}

\vspace{-20pt}

\subsection{Implementation details}
Our network is inspired by the standard Mask RCNN with two new attached branches as described in Figure 1. The initial learning rate is set to 0.001 and reduced in epochs 45 and 49 with a decay factor 0.1. The SGD with momentum is used as the optimizer and momentum is set to 0.9, while weight decay is set to 0.0001. We train the model with 50 epochs, with the loss function provided at equation(5), where $\beta$ is set to 0.01 according to our experience, note that $\beta$ cannot be set too large to avoid the bad influence of the pseudo-labeling phase. The $\lambda$ in equation(1) is set to 1. The mAP@0.5 is used as evaluation metric when inferencing on the test dataset, mAP@0.5 means using an IoU threshold 0.5 to identify whether a predicted bounding box or mask is positive in the evaluation. Mask IoU is used for mAP@0.5 in our experiment unless noted. For active learning, the train dataset is used as the unlabeled dataset $\emph{D}_u$, we set the labeled budget $B$ to 3000 and $b$ is equal to 500 per cycle. For pseudo-labeling in semi-supervised learning, we set $\sigma_c=0.9$, $\sigma_b=0.9$ and $\sigma_m=0.8$ per cycle. The other parameters are same as the Mask RCNN~\cite{he2017mask},  the total parameters and computations of the model are 47.07 M and 132.54 GFLOPs. The experiments are done with 8 NVIDIA V100 GPUs. Note that each experiment was repeated three times, and the results were averaged as the final results. 

\begin{figure}
\centering
\includegraphics[width=4.8in]{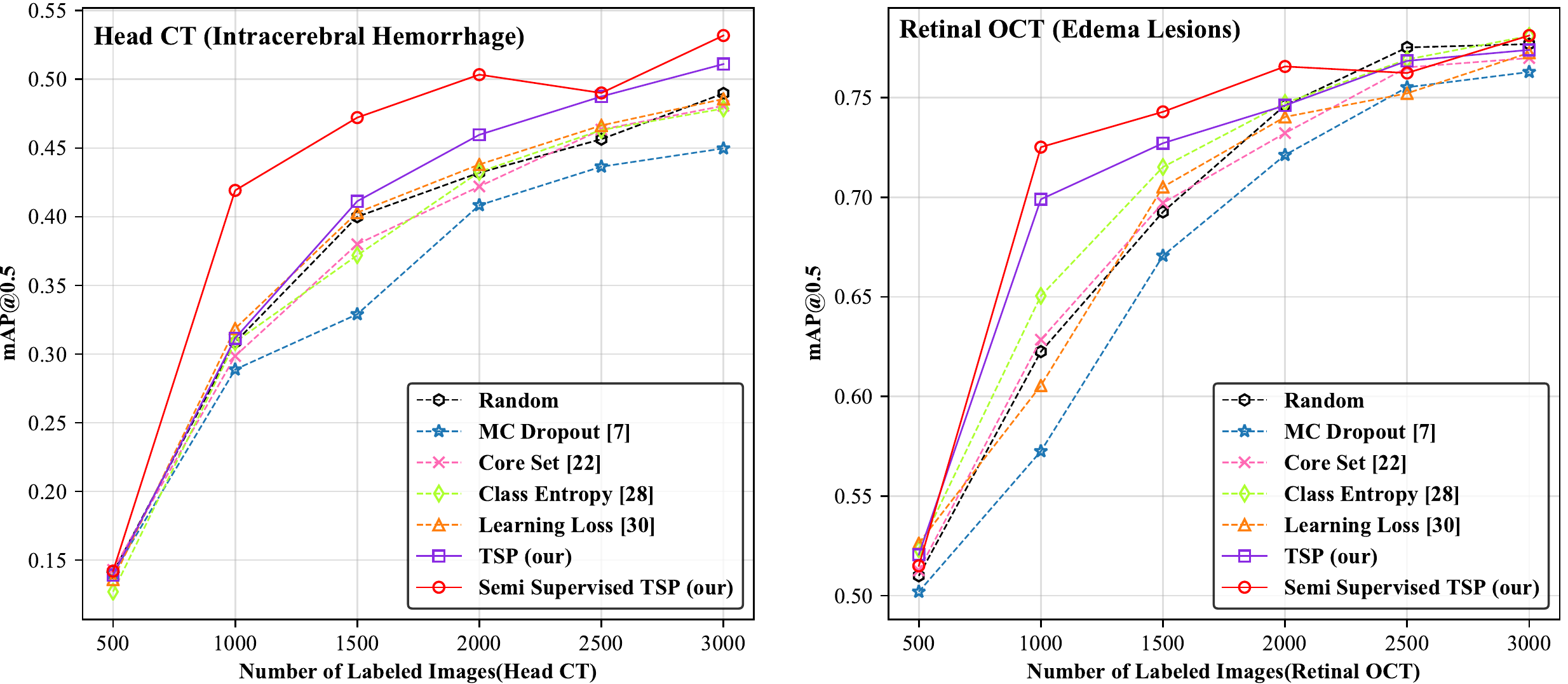}
\caption{mAP@0.5 on the test dataset of head CT dataset(left) and retinal OCT dataset(right) for instance segmentation with different active learning methods.} \label{fig2}

\vspace{-0.5cm}

\end{figure}

\section{Results and Discussions}
\noindent{\bf General Performance:} We compare our method with random baseline, MC dropout~\cite{gorriz2017cost}, Core set~\cite{sener2017active}, Class entropy~\cite{wang2016cost}, and Learning loss~\cite{yoo2019learning}, which are recent approaches for deep active learning. For random sampling, we randomly select $b$ samples from $\emph{D}_u$ per cycle and add them into $\emph{D}_{al}$ for training. For class entropy-based sampling, we compute the entropy of an image by averaging all entropy values from Softmax outputs corresponding to detection instances. We not only evaluate our proposed triplet scoring prediction strategy, then one step further, combine the strategy with an improved pseudo-labeling using our triplet scores for instance segmentation.

Figure 2 shows the mAP@0.5 curves for different uncertainty strategies on two datasets. In terms of entropy-based sampling, there is no obvious improvement comparing with random-based sampling. In particular, for CT datasets, the learning loss method perform better than random-based method and achieve about 0.9 mAP gain comparing with the random-based sampling at the last cycle. The performance of our proposed triplet scoring prediction (TSP) strategy surpass over the learning loss method with about 1.0 mAP gain at the last cycle, which indicated that our proposed method can suggest more informative samples for the model. The performance of active learning has obtained significant gain by semi-supervised TSP, which combines the proposed uncertainty strategy and improved pseudo-labeling for instance segmentation, it significantly performs better than other uncertainty strategies in all cycles
and achieve the maximum gain of 10 mAP comparing with the random-based sampling at the second cycle. Thus, as expected, since samples contribute to learning differently, thus prioritizing data for labeling pushed performance as expected, results demonstrate that such an scoring prediction measure enables informative samples ranking and taking full advantage of the unlabeled images. 

For comparison with non-interactive methods, in ICH segmentation task, our method using 1500 CT images yields mAP of 0.4869 in 5.2 hours, by adding 500 per cycle, meanwhile, the baseline directly trained with 3000 images yields similar mAP 0.4892 in 4.8 hours. More importantly, labeling a CT image takes roughly 3-5 minutes and costs 7-10 dollars for an annotation expert, consequently, besides training cost, our method reduces annotation cost from 3000 to 1500 images, approximately saving at least 4500 minutes and 10500 dollars.

\begin{figure}
\centering
\includegraphics[width=5in]{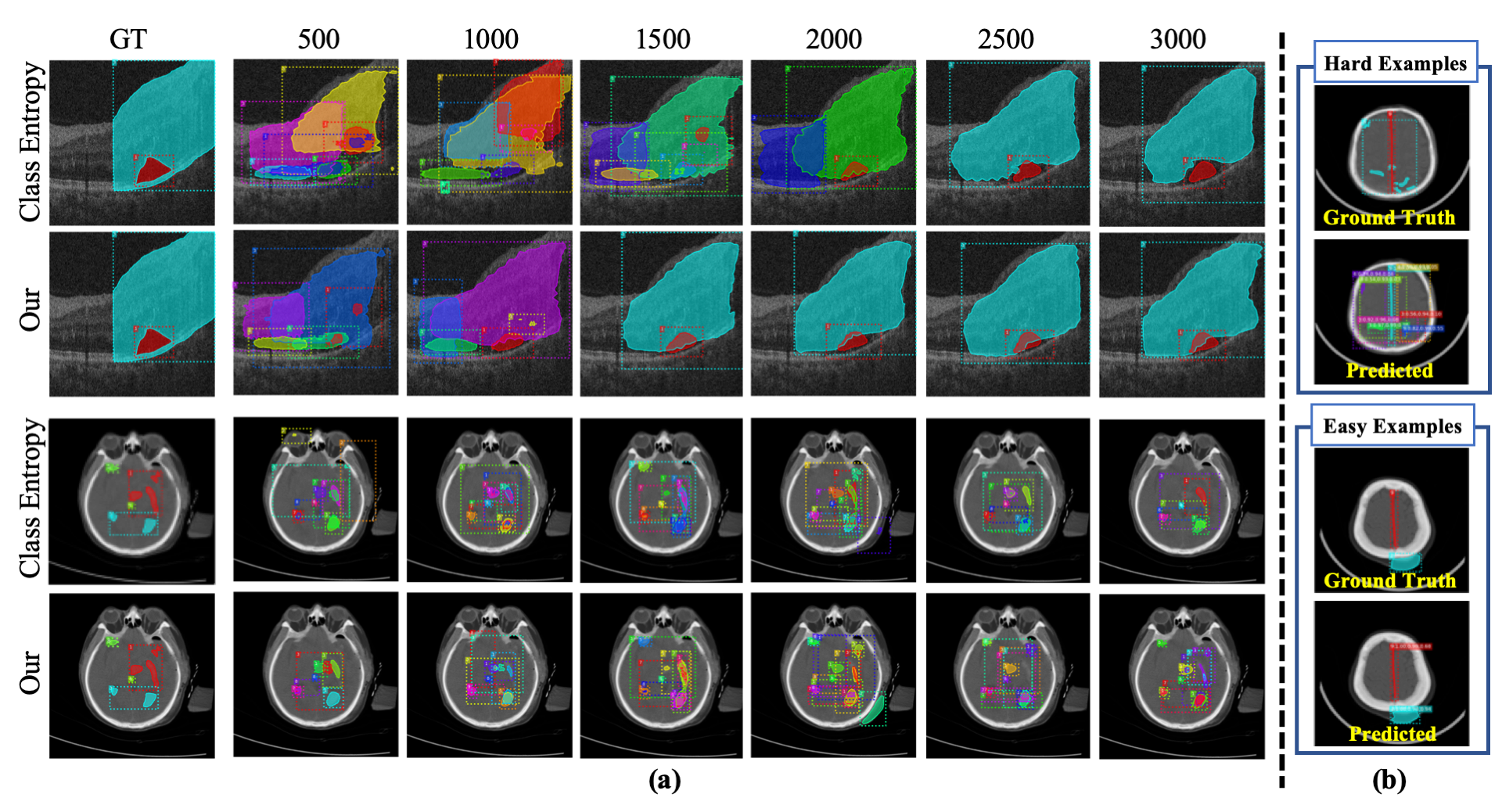}
\caption{(a) Qualitative instance segmentation results on at each cycle comparing the Entropy and our method (upper 2 columns for edema lesions segmentation in OCT, lower 2 columns for intracranial hemorrhage segmentation in CT). The column headings indicate the budget used to train the corresponding model. (b) Informative hard examples and relatively easy examples from CT dataset, according to our proposed triplet scoring strategy. } \label{fig3}
\vspace{-0.5cm}

\end{figure}


\noindent{\bf Discussion:} As shown in Figure 3(a), the result indicates that the proposed semi-supervised active learning method can achieve better performance in earlier rounds with less annotation, compared with commonly used methods like class entropy. Thus, not only the information in uncertain samples is helpful for saving label costs, but also the certain samples. Exploring both actively queried uncertain datasets and unlabeled certain datasets can offer a better representation of the underlying sampling space, other than just focusing on the most uncertain samples. Besides, the noise in the pseudo-labeling data will disturb performance, the balance weight beta should reduce as $\emph{D}_{al}$ increasing.Therefore, the pseudo-labeling is applied for excavating the informative in the unlabeled dataset, of course, a novel pseudo-labeling strategy and a proper loss design have been proposed for reducing the sampling bias of pseudo-labeling in this paper. 

Interestingly, an important insight observed from the results is that instance-level sampling is a more reasonable way than image-level sampling for active learning on instance segmentation task with higher labeling costs. Moreover, our results also indicate that when adapting AL methods, there should always be a direct combination of AL methods and SSL methods, it consolidates the robustness and replicability of AL sample selection.


\begin{figure}
\centering
\includegraphics[width=5in]{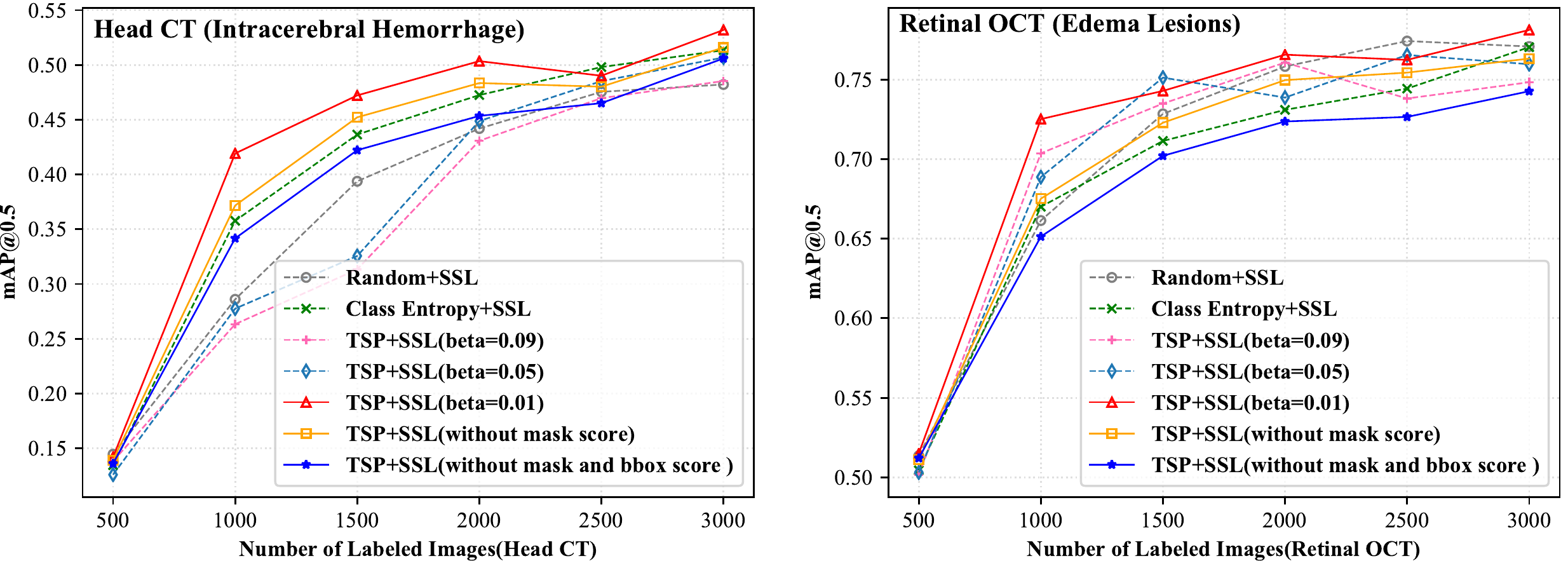}
\caption{Ablation study. The effectiveness verification on head CT dataset(left) and retinal OCT dataset(right). (1) the triplet scoring and pseudo labeling components: line 1, 2, 5; (2) the losses balancing coefficient $\beta$ between supervised and pseudo-labeling phase: line 3, 4, 5; (3) the composition of triplet score (removing box/mask score branch): solid lines 5, 6, 7.} \label{fig4}

\vspace{-0.5cm}

\end{figure}

Typical hard examples and easy examples are visualized in Figure 3(b) according to our proposed uncertainty ranking strategy, the method substantially provides high-level "explanations" for better informative sampling. The predicted result of each instance on the images have the format: class id: classification score, bounding box score, mask score. Hard examples with lower scores, appears when the predicted results contains too much false-positive instances or the mask score of instance is too low. For easy examples with higher uncertainty scores, the difference between the predicted results and the ground truth is very small, making it possible to be pseudo-labeled for training directly. 

Given that medical image datasets in similar reported works are usually constrained in limited size and single task, however, we evaluate the proposed method on substantially larger datasets than earlier papers on this topic. It is worth noting that our proposed method can be generalized to the natural images. This is because our network is based on Mask-RCNN, which achieves a great performance on natural images, meanwhile we can also obtain triplet scoring predictions of each instance for a natural image accordingly, then perform active learning and pseudo labeling via the triplet scoring predictions. In addition, we would add additional experiments on different imaging tasks like segmentation of histopathological images and natural images in our future work. 

\vspace{+0.3cm}

\noindent{\bf Ablation Study:} As shown in Figure 4, (1) firstly, we investigate the effectiveness of triplet scoring prediction and pseudo-labeling, as two key components in our method.  we replace triplet scoring components with Random baseline and Class Entropy, to assess the resulting difference in performance. As expected, results indicate that our method (red line with default $\beta$=0.01) outperforms Random baseline+SSL and commonly used Class Entropy+SSL. More importantly, it suggested that AL and SSL should be combined together in practice. 
(2) In addition, the losses balancing coefficient $\beta$ between training and pseudo-labeling is evaluated, the results indicates that strictly using smaller $\beta$=0.01 can better handle the pseudo-labeling noise than 0.05 and 0.09. 
(3) Furthermore, in the composition of triplet score, the class/bounding box/mask scores were evaluated in an ablation manner to show different kinds of effects. The line with rectangle marker (remove bbox score) and line with star marker (remove both bbox score and mask score)  in Figure 4 showed that, adding bbox score branch and mask score branch lead to gradual improvements of performance, and mask score contributes more significant performance improvement than bbox score and class score.


\vspace{-20pt}

\section{Conclusion}

In this paper, we propose a  semi-supervised active learning architecture to achieve significant savings in annotation effort required to train deep instance segmentation networks, it incorporates a novel scoring predictions strategy and progressive pseudo labeling, it push the envelope of performance with minimal labeled data in AL scenario, by exploiting an unlabeled data pool for instance segmentation model training. By comparative experiments on two image analysis tasks, we show that our method outperforms other active learning methods. We believe that our work could open up the possibilities in label-efficient active learning for instance segmentation.

\noindent{\bf Acknowledgements:} This work was supported by National Natural Science Foundation of China Grant No. 61806044.


\bibliography{egbib}

\end{document}